\documentclass[runningheads,a4paper]{llncs}

\usepackage{amssymb}
\usepackage{graphicx}
\usepackage{arydshln} 
\usepackage[hyphens]{url}
\usepackage{tikz}
\usetikzlibrary{arrows, petri, topaths}
\usepackage{tkz-berge}
\usepackage{float}

\urldef{\mailsa}\path|{gwohlg}@corp.ifmo.ru|    
\newcommand{\keywords}[1]{\par\addvspace\baselineskip
\noindent\keywordname\enspace\ignorespaces#1}

\begin{document}
\sloppy

\mainmatter  

\title{Using Word Embeddings for Visual Data Exploration with Ontodia and Wikidata}
\titlerunning{Using Word Embeddings for Visual Data Exploration}


\author{Gerhard Wohlgenannt\inst{1} \and Nikolay Klimov\inst{1} \and Dmitry Mouromtsev\inst{1} \and Daniil Razdyakonov\inst{2} \and Dmitry Pavlov\inst{2} \and Yury Emelyanov\inst{2}}



\institute{Intern.~Lab.~of Information Science and Semantic Technologies, ITMO University, St. Petersburg, Russia \url{http://en.ifmo.ru/en}
\and Vismart Ltd., St. Petersburg, Russia \url{https://vismart.biz}}

%
%

\maketitle

\begin{abstract}

One of the big challenges in Linked Data consumption is to create visual and natural language
interfaces to the data usable for non-technical users. Ontodia provides support for diagrammatic
data exploration, showcased in this publication in combination with the Wikidata dataset.
We present improvements to the natural language interface regarding exploring and querying
Linked Data entities. The method uses models of distributional semantics to find and rank entity properties
related to user input in Ontodia. Various word embedding types and model settings are evaluated,
and the results show that user experience in visual data exploration benefits from the proposed approach.

\end{abstract}

\setlength{\textfloatsep}{15pt}
\setlength{\intextsep}{15pt}
\section{Introduction}
\label{sec:intro}

The gigantic data source of Linked Data (LD) is accessible both by machines and humans.
Especially for end users, there are high barriers, such as finding relevant datasets,
understanding the schema, or being familiar with query languages such as SPARQL~\cite{Augenstein2013}.
One of the tools that provide an intuitive way to discover LD for non-technical users
is Ontodia\footnote{\url{http://www.ontodia.org}}. Ontodia is an open-source library for OWL and RDF diagramming and visual
exploration. In its current version, natural language (NL) search in the properties of given entities 
will only find properties exactly matching in the its labels. Here, we investigate a method to make
the search more flexible and abstracting users from the underlying data schemata by leveraging 
word embeddings to provide properties which are semantically related to a user query.
Using Wikidata\footnote{\url{https://www.wikidata.org}} as underlying dataset, we aim to i) investigate if word embeddings
are useful for the given problem, ii) evaluate which types of pre-trained embedding models, and which parameters, are best suited for the 
task, and iii) provide a prototype to demonstrate the benefits of the method.

We do not aim at full-fledged question answering over LD with NL to SPARQL transformation, 
but at improving the search functionality in diagrammatic LD exploration.





\section{Related Work}
\label{sec:related_work}

Query expansion for keyword queries is a classical problem in information retrieval.
A traditional way of keyword expansion is the use of dictionaries such as WordNet to find synonyms
or hypo- and hypernyms. This method suffers from sparse data regarding Named Entities and missing coverage of 
specialized domains.
In the Semantic Web field, eg.~Augenstein et al.~\cite{Augenstein2013} propose a method to map keywords
to LD resources by finding the properties that are related to semantic similarity between resources.
In contrast to our work, which searches in entity properties, Augenstein et al.~\cite{Augenstein2013} focus 
primarily on finding resources (entities).
Freitas et al.~\cite{Freitag2013} propose a complex system for querying heterogeneous, and distributed datasets, which
abstracts users from the underlying data schemata. The system combines entity search, a Wikipedia-based
semantic relatedness measure and spreading activation to answer NL queries.

Challenges and future directions in Question Answering on LD are presented in Shekarpour et al.~\cite{Shekarpour2016}.
The application of word embeddings and deep learning is listed prominently among the promising techniques for future investigation.
In line with this recommendation, we apply distributional semantics for the natural language query interface of Ontodia.
In general, word embeddings transform the vocabulary of a given corpus into a continuous low-dimensional vector space representation.
They have been successfully applied, for example, for word similarity computations, but also more complex natural language tasks~\cite{ghanny2016}.

\section{System Description}

\begin{figure*}[htb]
\centering
{\centering \resizebox*{0.73\textwidth}{!}{
\includegraphics{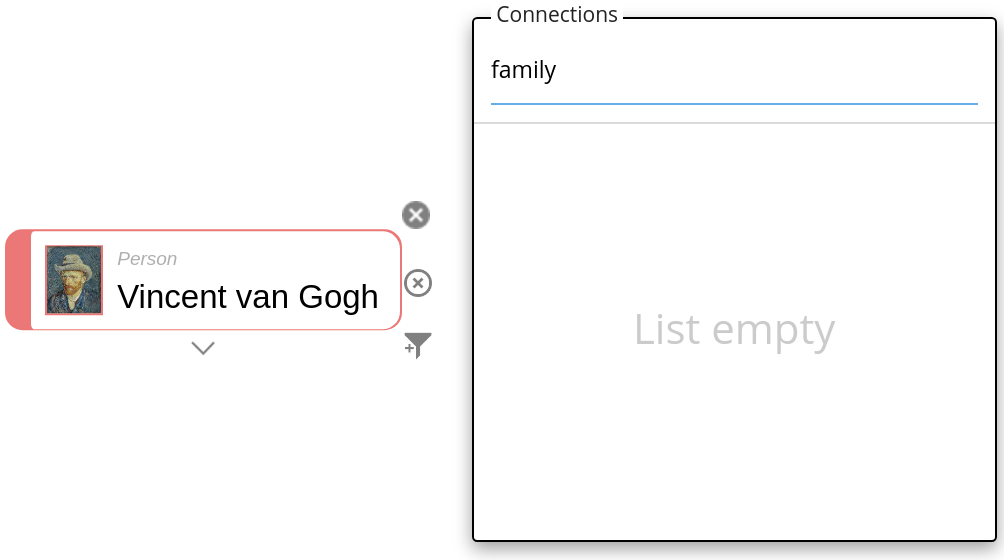}
}} \caption{\label{fig:sc1} Searching for ``family'' relations of entity \emph{Van Gogh} in the original system.}
\end{figure*}

The work presented in this paper extends Ontodia with improved search capabilities.
As mentioned, Ontodia is an open-source tool\footnote{\url{https://github.com/ontodia-org/ontodia}} 
for simple OWL and RDF visual data exploration. Ontodia is often integrated with 
metaphactory\footnote{\url{http://www.metaphacts.com/product}} as a semantic platform backend.
In a typical data exploration scenario, the user starts querying the dataset
at the system entry point\footnote{\url{https://wikidata.metaphacts.com/resource/Start}}.
At search result, the user can switch to using Ontodia to explore the data space.
In the current version, search in the connections of an entity only finds literal matches of
the search term in the property labels. This limits the ease-of-use with unfamiliar datasets.
E.g, when looking for family relations of entity \emph{Van Gogh}, the system will
not find any matching properties due to missing exact lexical matches, see Figure~\ref{fig:sc1}.

\label{sec:methods}

\begin{figure*}[htb]
\centering
{\centering \resizebox*{0.73\textwidth}{!}{
\includegraphics{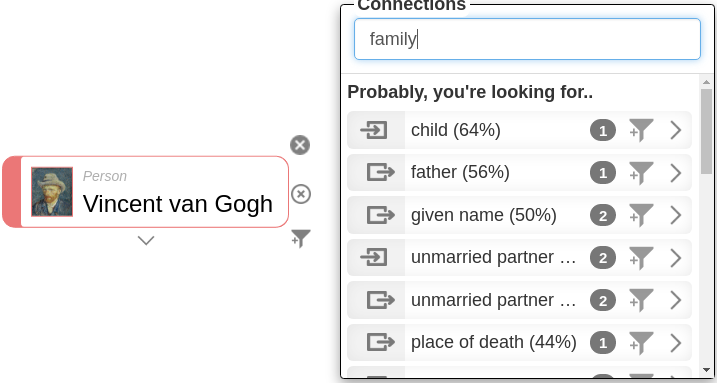}
}} \caption{\label{fig:sc2} Searching for ``family'' relations of entity \emph{Van Gogh} in the new prototype.}
\end{figure*}

The prototype presented here makes use of a) aliases for property labels defined in Wikidata, and it applies 
distributional semantics in the form of word embeddings to find suitable properties related to a user query.
Figure~\ref{fig:sc2} shows the results using the new search functionality, which are a combination of:
(i) exact matches of the input term in the property labels, (ii) exact matches in property aliases,
and (iii) related properties according to the word embedding model used, ordered descendingly by semantic similarity.

The updated search interface also allows for a new way of data exploration, where the user is interested in a certain topic,
for example \emph{family} or \emph{politics}, and can then explore all entity properties (connections) related to the topic.\\ 
The prototype described here is available at:\\
\url{http://ontodia-prop-suggest.apps.vismart.biz/wikidata.html}.

\subsection{The Method}

A central ingredient to the method is the word embedding model. 
The models were trained on a Wikipedia corpus -- and in some cases additional textual sources --
and contain continuous vector space representations of the words from the corpora 
which capture the distributional semantics of the words.

First, the Wikidata properties need to be added to the model vector space. 
For every property we split the property label (\texttt{rdf:label}) into a list of words,
and remove stopwords. The vector representation of a property is created as the vectorial sum of the words.
A variant of the system also includes the words from the property descriptions to create the property vectors.
At runtime, the same process is applied to the natural language user query provided in the search box. 
The query is split into single words, stopwords are removed, and the vectorial representation is the sum
of the query word vectors.
Finally, the system ranks the properties 
by cosine similarity between the query vector and all the property vectors to find the most relevant properties.

The method is simple and computationally efficient. In this publication, the focus is on the evaluation of the method,
and especially on comparing the performance of various types of word embeddings.

\subsection{Implementation}

The presented method and the accompanying code 
was implemented in Python and
can be found on GitHub\footnote{\url{https://github.com/gwohlgen/ontodia_search_properties}}. 
The main modules include a preprocessing phase, where the vectors for the Wikidata properties are 
constructed and persisted, the module to rank properties according to user input, and the tools for the evaluation of the system.
For integration with Ontodia, we created a webservice that takes the user input in JSON format,
computes the property rankings, and returns them in JSON format to Ontodia for display to the user.

\section{Evaluation}
\label{sec:evaluation}

First, this section describes aspects of evaluation setup like the Wikidata dataset, the gold standard data used,
system settings and the word embedding models. Then, a detailed presentation of the evaluation results, 
including a discussion of aspects like dataset quality and result interpretation, follow.

\subsection{Evaluation Setup}
\label{sec:eval_setup}

\subsubsection{Wikidata Dataset}

Wikidata is an open knowledge base, which can be exported and interlinked with other datasets on the Linked Data web.
Wikidata is the central data storage for projects like Wikipedia.
The dataset currently includes around 28 million items, and, more relevant for this work,
there are 3323 properties defined to describe and connect the entities.  
The properties have labels for various languages, and aliases (called ``also known as'') for many of the labels. 
We focus on English language labels, for which currently 4603 aliases are defined. 
Additionally, properties usually have a short textual description, which we also use in our method to create
property representations.

\subsubsection{Gold Standard Dataset}

The aliases manually defined in Wikidata are an obvious source to be used as a gold standard dataset to evaluate our method. 
For this purpose, any of the 4603 English language aliases is used as an query term, 
and the system suggests a ranking of properties similar to the term. 
1736 of 3323 properties actually have aliases defined. 

%
Despite the varying quality of aliases (details in the \emph{Discussion section}), we decided to use them as a gold standard dataset. 
Eventually the proposed method can even be applied to help detect questionable alias definitions in the future.



\subsubsection{System Settings}

In the evaluations, we experimented with various system settings and word embedding models.
The types of word embedding models are described below, the most important system settings include:

\begin{itemize}
\item \textbf{Use description text (Boolean):} For creating the representations of properties in vector space, 
we compared the results of using only the words from the property labels versus words from property labels and description texts.
\item \textbf{Dimensions of vector model:} Some predefined vector models are available with different numbers of vector dimensions
(for example 50 vs. 100 vs. 300 dimensions).
A lower number of dimensions makes the model more computationally efficient, but it may loose semantic nuances.
\item \textbf{Number of words in the model:} In the pre-trained models the word vectors are ordered descendingly
by word frequency in the training corpus. Big models with hundreds of thousands of vectors occupy a lot of memory and take a long time load. 
Therefore, we compared the performance of models with 300.000 words with smaller models with the 10.000 most frequent words.
\end{itemize}

\subsubsection{Word Embeddings}

One of the main goals was to evaluate which of the pre-trained word embedding models is best
suited for the task at hand. The pre-trained models available are not trained on exactly the same corpus, 
but all include English Wikipedia. The following word embedding types were evaluated:

\begin{itemize}

    \item \textbf{fastText}:
        FastText~\cite{bojanowski2016} is an extension of the original Word2vec~\cite{mikolov2013w2v} model which uses sub-word information.
        Words are represented as bag of character n-grams. FastText generates better word embeddings for rare words,
        and takes morphological information into account.
        Here, we applied a model trained on Wikipedia 2016\footnote{\url{https://github.com/facebookresearch/fastText/blob/master/pretrained-vectors.md}}.
        Two variants were compared, a model with 300.000 words, and a small model with only the 10.000 most frequent words.

    \item \textbf{GloVe}:
        GloVe\cite{glove} factors the logarithm of the co-occurrence matrix that reflects the position of the context words in the word window.
        We used a model pre-trained on a Wikipedia 2014 and Gigaword 5 corpus (6B tokens)\footnote{\url{https://github.com/stanfordnlp/GloVe}}.
        Variants include combinations of models with 300, 100 or 50 dimensions, and 300.000 versus only 10.000 word vectors.

    \item \textbf{LexVec}:
        LexVec~\cite{SalleIV16a} is a word embedding method which factorizes PPMI matrices and combines characteristics of techniques like Word2vec and GloVe.
        LexVec performs well on word similarity and semantic analogy tasks, but struggles on syntactic analogies.
        The model used was trained on a 7B token corpus of English Wikipedia 2015 and NewsCrawl\footnote{\url{https://github.com/alexandres/lexvec}}.
        Again, we evaluated variants of 300.000 versus 10.000 word vectors.


\end{itemize}

\subsection{Evaluation Results}

In the main evaluation which aims to judge the suitability of various word embedding types we experiment
with different models and settings. As stated, the task is as follows: for any of the aliases defined for Wikidata properties,
we create a ranking of related properties. The word vectors of the alias words are compared to the vectors representing the properties.
Every alias is compared to all 3323 properties, which is much harder than the real-world task of searching only in the properties of a given
entity. The later task is evaluated in the next section.

Table~\ref{tab:main_eval} presents an overview of the results. 
Column one states the embedding model type and the settings, namely the model size (either 300.000 or 10.000 words), 
and the dimensions of the vectors. 
The metrics \emph{Top-N} reflect the ratio of system suggestions, where the correct property is in the \emph{Top-N} 
of the generated ranking.
MRR is the well-known \emph{mean reciprocal rank}.
The lower part of the table includes some results for models which only use the words from the property label 
to create the property vectors, but not from the description text (\emph{WO-D}).

\begin{table}
\begin{center}
\begin{tabular}[c]{|c|c|c|c|c|} \hline
                                          & ~\textbf{Top 1}~~& ~~\textbf{Top 3}~~~& ~~~\textbf{Top 10}~ & ~~~\textbf{MRR}~ \\ \hline\hline
    \textbf{fastText} 300.000 / 300d      & \textbf{38.12\%} & \textbf{55.13\%}   & \textbf{70.22\%}    & \textbf{0.493}   \\ 
    \textbf{fastText}  10.000 / 300d      &        31.49\%   &       48.59\%      &       66.29\%       &      0.432       \\ 

    \textbf{GloVe}    300.000 / 300d      &        36.33\%   &       52.82\%      &       66.55\%       &     0.469        \\ 
    \textbf{GloVe}     10.000 / 300d      &        30.54\%   &       45.90\%      &       61.78\%       &     0.411        \\ 
    \textbf{GloVe}    300.000 / 100d      &        33.24\%   &       47.85\%      &       61.94\%       &     0.429        \\ 
    \textbf{GloVe}     10.000 / 100d      &        28.39\%   &       43.21\%      &       58.97\%       &     0.386        \\ 
    \textbf{GloVe}    300.000 /  50d      &        27.97\%   &       41.88\%      &       56.58\%       &     0.376        \\ 
    \textbf{GloVe}     10.000 /  50d      &        24.42\%   &       38.51\%      &       54.24\%       &     0.344        \\ 


    \textbf{LexVec}   300.000 / 300d      &        37.21\%   &       53.45\%      &       67.99\%       &     0.479        \\ 
    \textbf{LexVec}    10.000 / 300d      &        30.59\%   &       46.03\%      &       62.55\%       &     0.411        \\ \hline

    \textbf{fastText WO-D} 300.000 / 300d &        36.10\%   &       51.94\%      &       65.25\%       &     0.464        \\ 
    \textbf{fastText WO-D}  10.000 / 300d &        29.99\%   &       46.83\%      &       60.74\%       &     0.407        \\ 
    \textbf{GloVe WO-D}    300.000 / 300d &        34.21\%   &       48.88\%      &        60.76\%      &     0.437        \\ \hline \hline 
\end{tabular}
\caption{Evaluation of word embedding models and settings on aligning Wikidata aliases to the corresponding property.}

\label{tab:main_eval}
\end{center}
\end{table}

The fastText model with 300.000 word vectors and 300 vector dimensions performs best over all metrics. 
We also experimented with a bigger fastText model with around 2.5m word vectors, but those additional rare words just increased memory consumption,
the performance stayed almost the same. 
On the other hand, it is evident that reducing the model size to 10.000 words affects performance
negatively. Over all model types reducing model size from 300.000 to 10.000 words led to a sharp drop in accuracy.
Regarding model types, fastText is best suited for the task, followed by LexVec, and lastly GloVe. 
As seen in the last part of the table, using the words from the description text to represent property vectors is helpful. 
Finally, using fine grained word representations with larger vectors (50 versus 100 versus 300 dimensions) has a strong positive
effect.

\subsubsection{Property Search for Single Entities}
\label{sec:single_entity}
In the evaluations above, we measure the accuracy for matching aliases against all the 3323 properties in Wikidata.
However, in an interactive scenario of visual data exploration with Ontodia, the user query is typically restricted to the properties defined
for a specific entity. This scenario was simulated and evaluated by randomly choosing 1150 entities from the Wikidata dataset,
and performing the evaluation with the their properties and aliases.
In total, about $85\%$ of the properties had one or more aliases defined, with an average of $5.9$ aliases per property.
Table~\ref{tab:rand_ent} presents the evaluation using the fastText and LexVec models on the task of finding the corresponding 
entity property for all aliases defined for an entity. 




\begin{table}
\begin{center}
\begin{tabular}[c]{|c|c|c|c|c|} \hline
                                          & ~\textbf{Top 1}~~& ~~\textbf{Top 3}~~~& ~~~\textbf{Top 10}~ & ~~~\textbf{MRR}~ \\ \hline\hline
    \textbf{fastText}      &   68.63\%        &  84.75\%           &  94.59\%            & 0.78 \\
    \textbf{LexVec}        &   62.08\%        &  80.93\%           &  93.18\%            & 0.73 \\ \hline

\end{tabular}
\caption{Evaluation of system accuracy for matching aliases with properties of randomly-picked entities.}
\label{tab:rand_ent}
\end{center}
\end{table}

Again, fastText outperforms the LexVec embeddings. When ranking the entity properties for the alias term by similarity,
in over $70\%$ of cases the first ranked property is correct with respect to the gold standard. For the Top-3 results, the number is
$87.49\%$, and the MRR is $0.80$. The results make us confident that the new search feature has a very positive impact
on user experience. 
The runtime of a query is typically under $10ms$ -- well-suited for interactive systems.

\subsection{Discussion}

\subsubsection{Dataset Quality}

During the evaluation and the inspection of the results we found various issues with Wikidata dataset quality, 
which (i) explain part of the misclassification of the method, 
and (ii) provide hints on improving dataset quality, esp.~the quality of aliases. 
First of all, in 14 cases the alias was exactly the same term as the property label. More interestingly, 
many aliases are not proper synonyms. For example, property \emph{P582} with label ``end time'', has alias such as ``divorced'', or simply ``to''.
Or, \emph{P150} with label ``contains administrative territorial entity'', has aliases such as ``divides into'', ``contains'', ``has villages'' 
-- some of which make it hard for the system to link to the correct property.

\subsection{System Performance}

The experiments summarized in Table~\ref{tab:main_eval} indicate that the fastText algorithm is best suited 
for the task, followed by LexVec. System configuration, especially the model vocabulary size 
and the number of vector dimensions are crucial for system performance, and should only be compromised 
if decreasing memory footprint is inevitable.
Furthermore, including the property descriptions in the vector provides better property representations.
In our real-world use case (Section~\ref{sec:single_entity}) the method demonstrates sufficient performance to improve user experience.

Regarding computational performance, using Python and the gensim library, the fastText model with 300.000 vectors and 300 dimensions 
consumes ca.~650M of memory, a 10.000 words model requires 130M.
The runtime for a query against all 3323 properties is around 300ms, for the interactive use-case query time is usually below $10ms$.

\section{Conclusions}
\label{sec:concl}


In this publication we present a method for simple and powerful search in entity properties of Linked Data 
using natural language. A prototype of the method is integrated into the Ontodia tool using Wikidata as data source.
The method applies models of distributed semantics to find properties related to user input. 
The contributions include (i) the presentation of a method for searching in Linked Data
which applies word embeddings to the given task in an efficient way, 
(ii) an extensive evaluation of various types of word embedding models and parameters such as model size and dimensionality against a gold standard, 
(iii) the provision of the implementation and an online prototype.
%
In future work we will apply the presented approach to other datasets, and investigate the integration with
more powerful question answering for Linked Data techniques.


\section{Acknowledgments}
This work was supported by the Government of the Russian Federation (Grant 074-U01) through the ITMO Fellowship and Professorship Program.

\bibliographystyle{splncs03}
{\small \bibliography{here2}}


\end{document}